\documentclass[10pt,twocolumn,letterpaper]{article}

\usepackage{iccv}              % To produce the CAMERA-READY version
\usepackage[ruled,vlined]{algorithm2e}
\usepackage{amsmath}  % 支持数学符号
\usepackage{amssymb}  % 支持更多数学符号
\usepackage{algpseudocode} 
\usepackage{multirow}
\usepackage{booktabs}
\usepackage{tcolorbox}
\usepackage{float}
\usepackage{threeparttable}
\usepackage{stfloats}
\usepackage{balance}
\usepackage{graphicx}
\usepackage{caption}
\usepackage{lipsum}
\usepackage{setspace} % 用于设置行间距
\usepackage{multicol}
\usepackage{array,tabularx}
\usepackage{pifont}
\usepackage{tabularray} 
\usepackage{subcaption}

\definecolor{iccvblue}{rgb}{0.21,0.49,0.74}
\usepackage[pagebackref,breaklinks,colorlinks,allcolors=iccvblue]{hyperref}

\def\paperID{} % *** Enter the Paper ID here
\def\confName{}
\def\confYear{}

\title{Optimization of Low-Latency Spiking Neural Networks Utilizing Historical Dynamics of Refractory Periods}

\author{Liying Tao\\
Institute of Microelectronics of the Chinese Academy of Sciences\\
University of Chinese Academy of Sciences\\
{\tt\small taoliying@ime.ac.cn}
\and
Zonglin Yang\\
Nanjing Institute of Intelligent Technology\\
{\tt\small yzl@niit.ac.cn}
\and
Delong Shang\\
Institute of Microelectronics of the Chinese Academy of Sciences\\
Nanjing Institute of Intelligent Technology\\
{\tt\small shangdelong@ime.ac.cn}
}

\begin{document}
\maketitle
\begin{abstract}

The refractory period controls neuron spike firing rate, crucial for network stability and noise resistance. With advancements in spiking neural network (SNN) training methods, low-latency SNN applications have expanded. In low-latency SNNs, shorter simulation steps render traditional refractory mechanisms, which rely on empirical distributions or spike firing rates, less effective. However, omitting the refractory period amplifies the risk of neuron over-activation and reduces the system's robustness to noise.

To address this challenge, we propose a historical dynamic refractory period (HDRP) model that leverages membrane potential derivative with historical refractory periods to estimate an initial refractory period and dynamically adjust its duration. Additionally, we propose a threshold-dependent refractory kernel to mitigate excessive neuron state accumulation. Our approach retains the binary characteristics of SNNs while enhancing both noise resistance and overall performance. Experimental results show that HDRP-SNN significantly reduces redundant spikes compared to traditional SNNs, and achieves state-of-the-art (SOTA) accuracy both on static datasets and neuromorphic datasets. Moreover, HDRP-SNN outperforms artificial neural networks (ANNs) and traditional SNNs in noise resistance, highlighting the crucial role of the HDRP mechanism in enhancing the performance of low-latency SNNs. 
\end{abstract}    
\begin{figure}[t]
	\centering
	\begin{minipage}[c]{0.48\textwidth}
		\centering
		\includegraphics[width=\textwidth]{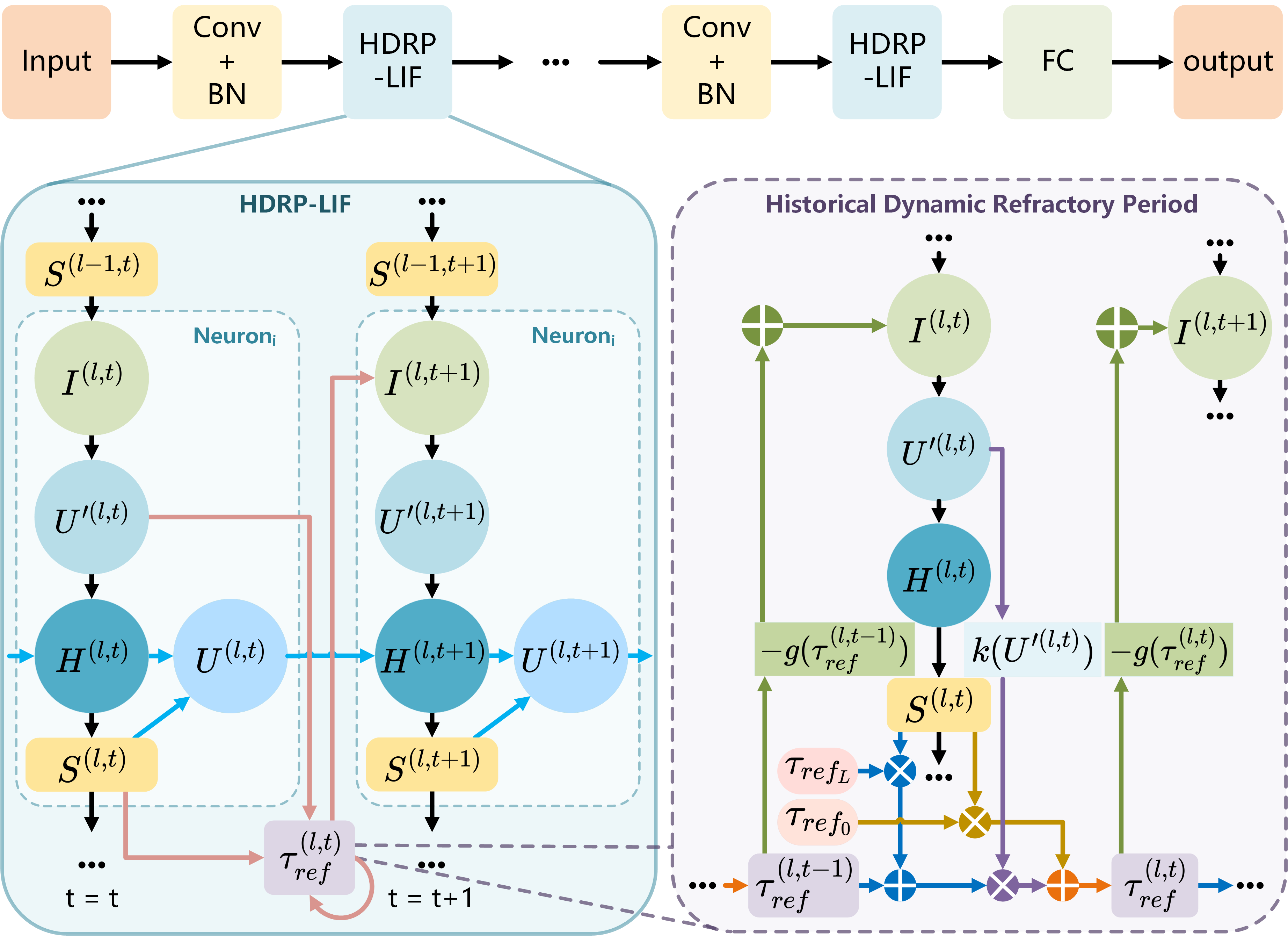}
		\caption{\textbf{HDRP-SNN architecture.} The internal state updates of the HDRP-LIF model are performed locally, where the refractory time constant $\tau_{ref}^{(l,t)} $ is modulated by both the previous time step $\tau_{ref}^{(l,t-1)}$ and the derivative of the current membrane potential. Additionally, the refractory kernel Inhibits the synaptic current at the next time step, effectively regulating the firing rate of the neurons. The HDRP-LIF can be seamlessly integrated into existing SNN frameworks while preserving the spike transmission characteristics of SNNs.}
        \label{fig:1}
	\end{minipage} 
    \vspace{-0.4cm}
\end{figure}

\section{Introduction}
\label{sec:intro}

SNNs differ fundamentally from ANNs in how they transmit information and perform computations \cite{guo2023direct}. ANNs rely on matrix operations and activation functions, updating neuron states through synchronized processing. In contrast, SNNs use discrete spikes and event-driven updates, eliminating the need for global synchronization \cite{bhattacharjee2024snns}. This decentralized computation allows each neuron to operate independently, enhancing efficiency in mimicking biological processes and achieving superior energy efficiency and computational potential \cite{li2022bsnn}. Neuromorphic hardware further capitalizes on asynchronous parallel processing, improving overall performance and reducing energy consumption \cite{angelidis2021spiking,young2019review,rathi2023exploring}. However, achieving comparable performance to ANNs while retaining energy efficiency remains a key challenge. 
% Recent advancements, such as surrogate gradient training, have improved SNNs’ accuracy and low-latency performance 
Recent advancements, such as surrogate gradient training, have significantly improved the accuracy and low-latency performance of SNNs
\cite{meng2023towards,taylor2024addressing,deng2023surrogate}. Additionally, biologically inspired models, which incorporate more realistic neuron dynamics and adaptive mechanisms, have further boosted SNN performance.\cite{xu2023biologically,geng2023hosnn}.

In biological neural systems, when neurons become excessively active, the spike signals they produce tend to obscure authentic, meaningful information, thereby reducing the overall efficiency of information processing within the neural network. Such over-activation can even lead to erroneous or chaotic signal transmission. Previous studies have demonstrated that excessive neuronal activation not only induces output saturation but may also disrupt local network synchrony and functional differentiation, a phenomenon particularly evident in complex dynamical systems \cite{koch2004biophysics, izhikevich2007dynamical}. Conversely, moderate inhibition helps maintain neurons at an optimal activity level, preventing the network-wide chaos and signal interference associated with hyperactivation, thus ensuring precise and stable information transfer. Additionally, appropriate inhibitory mechanisms enhance sensitivity to genuine, useful signals, ultimately improving the overall efficiency and responsiveness of the neural network \cite{buzsaki2006rhythms, isaacson2011inhibition}.

The refractory period plays a critical role in neural signal transmission within biological neurons \cite{song2017modeling}. Following spike generation, neurons enter either an absolute or relative refractory period \cite{de2024biophysics}, during which neuronal activity is suppressed either completely or partially. Typically, the refractory period duration is determined empirically and influenced by the neuron's firing frequency \cite{monk2016model,vardi2021significant,ratanov2020mean,bellec2020solution}. However, to simplify modeling, current low-latency SNN frameworks often neglect refractory mechanisms. Although this reduces computational complexity, it simultaneously increases the risk of over-activation in Leaky Integrate-and-Fire (LIF) neurons, leading to increased noise. The refractory period helps mitigate such noise by inhibiting spurious spikes, thus maintaining system stability and enhancing network efficiency and information processing quality \cite{avissar2013refractoriness}.

This phenomenon is particularly significant in low-latency SNNs, where the simulation timestep is relatively short and spike frequencies lack statistical reliability, complicating the accurate assessment of neuronal states. Spike emission in SNNs is modulated jointly by neuronal parameters and synaptic currents, making a fixed absolute refractory period insufficient for capturing the complexities of neural dynamics. Neuronal inputs are inherently time-varying; therefore, determining the refractory period based solely on instantaneous neuron states may result in inaccurate estimations and undermine the effectiveness of noise suppression. Consequently, for low-latency SNNs, the refractory period should be more precisely determined by incorporating historical spike frequencies, current neuronal states, and their initial conditions.

This paper proposes a HDRP estimation model. The model estimates the initial refractory period using the derivative of the membrane potential and historical spike firing rate, while dynamically adjusting the refractory period to adapt to neuron behavior in changing environments. This approach ensures that the refractory period reflects both frequency-dependent and strong input-driven inhibitory effects, aligning with the physiological characteristics of biological neurons. Additionally, a threshold-dependent refractory kernel is proposed such that the inhibition strength increases with the duration of the refractory period, effectively suppressing noise spikes while preserving rapid responsiveness to valuable inputs. The combined design of dynamic refractory periods and the refractory kernel enhances the robustness and computational efficiency of SNNs in processing complex spatiotemporal data, while preserving binary characteristics and facilitating low-latency, brain-inspired computing.

Contributions of this paper:
\begin{enumerate}

    \item We propose a novel historical dynamic refractory period (HDRP) model with a threshold-dependent refractory kernel for low-latency SNNs. Our unified time scale dynamically adjusts the refractory period and inhibition strength using the membrane potential derivative and historical firing rate, allowing HDRP-LIF to integrate with traditional LIF-SNNs without losing their binary nature.

    \item We demonstrate HDRP’s superior noise robustness by incorporating it into ResNet18 on CIFAR10 under varying Gaussian noise levels, where HDRP-SNN consistently outperforms conventional neuron models.

    \item We evaluate HDRP-SNN’s computational efficiency and overall performance on static datasets (CIFAR10, CIFAR100, ImageNet) and a neuromorphic dataset (DVS-CIFAR10). Extensive ablation studies confirm that HDRP enhances SNNs’ representational capacity, reduces accumulation operations, and achieves state-of-the-art or ANN-comparable performance while maintaining low energy consumption.

\end{enumerate}

\section{Related Works}

With advancements in SNN training, traditional challenges such as long simulation times and non-differentiability have been mitigated. Researchers have developed methods for directly training high-accuracy, low-latency SNNs, including gradient decomposition, removal of non-essential components to reduce training costs \cite{meng2023towards,wang2023ssf}, neuron parameter adjustments \cite{guo2023membrane,huang2024clif}, shortcut connections for mitigating gradient vanishing \cite{guo2024take}, and online learning \cite{yu2024advancing}. In SNN research, introducing a refractory period helps control excessive membrane potential accumulation and reduce spike firing rates, which achieves noise regularization \cite{berry1997refractoriness,zhu2020retina} and enhancing temporal encoding \cite{song2017modeling}. Moreover, the refractory period supports dynamic pattern formation, facilitating the generation of spatiotemporal patterns \cite{gong2013spatiotemporal}. However, its direct use in low-latency SNNs with direct training can destabilize network performance \cite{ikegawa2022rethinking,lin2024spiking}, especially under shorter simulation steps, which limit the effectiveness of traditional refractory period mechanisms.

Traditionally, the refractory period includes absolute and relative refractory periods which are typically set based on empirical values\cite{de2024biophysics,zhang2021self}. During the absolute refractory period, neurons cannot fire as membrane potential remains constant \cite{de2024biophysics}. The fixed refractory period, however, may limit the adaptability and overall performance of neurons when responding to variable inputs. In the relative refractory period, inhibition intensity decays exponentially \cite{de2024biophysics,menesse2024information}, allowing neurons to spike with stronger stimuli for a gentler modulation of firing rate.

To better control spike firing rate, researchers have explored implicit methods resembling a refractory period, such as raising firing thresholds based on frequency \cite{huang2024clif,shen2024efficient} and resetting states to extend membrane accumulation time \cite{yao2022glif}. These approaches suggest that inhibiting spikes could enhance the expressiveness of the network. Recent studies also suggest dynamically adjusting refractory period length based on cumulative spike count and fixed decay constants \cite{higuchi2024balanced}. 

However, in low-latency SNNs with limited simulation time, relying solely on spike count is insufficient for accurately estimating refractory requirements, and fixed decay rates lack adaptability. To address these limitations, we proposes the HDRP model and a threshold-dependent refractory kernel.

\section{Preliminary}

\subsection{Vanilla LIF Neuron Model}
The vanilla LIF neuron simulates the temporal computation of neurons by accumulating membrane potential and firing spikes upon reaching a threshold. Its differential equation is:

\begin{equation}
    \tau_m \frac{dU(t)}{dt} = - (U(t) - U_{\text{rest}}) + R_m I(t) \label
    {eq1}
\end{equation}
where $\tau_m$ is the time constant, $U(t)$ is the membrane potential of the neuron, $R_m$ is the membrane resistance, $U_{\text{rest}}$ is the resting potential, and $I(t)$ is the synaptic current.

To efficiently apply the LIF neuron in SNNs, we can use the explicit Euler method to discretize Equation (\ref{eq1}), transforming the continuous-time differential equation into a discrete form.
\begin{equation}
    U^{t+1} = \frac{1}{\tau_m}U^t + U_{\text{rest}}+ R_m I^t
    \label{eq2}
\end{equation}

\subsection{Loss Function}

In SNNs, we use a combination of cross-entropy loss and L2 regularization to optimize the model parameters. The cross-entropy loss ensures accurate classification, while L2 regularization helps prevent overfitting by penalizing large weights. The combined loss function is formulated as:

\begin{equation}
    L_\text{total} = \text{CrossEntropy}(\hat{y}(t), y(t)) + \frac{\lambda}{2} \sum_{i} \theta_i^2 \label{eq3}
\end{equation}
where $ \hat{y}(t) $ represents the model’s predicted output at time $ t $, while $ y(t) $ is the target label. $ \theta_i $ denotes the trainable weights of the model, $ \lambda $ is the coefficient for L2 regularization, and $ \gamma $ controls the gradient smoothness in the surrogate function.

\section{Method}

\subsection{Bounds on the Membrane Potential Derivative}
\label{subsec:4-1bounds}

Assume the SNN model is defined as follows::
\begin{equation}
    \begin{split}
        \hat{\mathbf{y}}(t) &= f\left( \mathbf{W}^{(l)} f\left( \mathbf{W}^{(l-1)} \cdots f\left(  \mathbf{W}^{(1)} \mathbf{x}(t)  \right. \right. \right. \\
        &\quad   \left. \left. \left. + \mathbf{b}^{(1)}\right) \cdots+ \mathbf{b}^{(l-1)} \right) + \mathbf{b}^{(l)} \right)
    \end{split}
    \label{eq4}
\end{equation}
where $f(\cdot)$ denotes the neuron model. $\mathbf{W}^{(l)}$ and $\mathbf{b}^{(l)}$ represents the weight matrix and the bias term of the $l$-th layer, respectively. $\mathbf{x}(t)$ is the input data at time $t$.

Under the constraint of the loss function in Equation (\ref{eq3}), the weight update rule for each layer $ \mathbf{W}^{(l)} $ is given by:
\begin{equation}
    \mathbf{W}^{(l, n+1)} = \mathbf{W}^{(l, n)} - \eta \nabla_{\mathbf{W}^{(l)}} L_{\text{CE}} - \eta \lambda \mathbf{W}^{(l, n)} 
    \label{eq5}
\end{equation}

where $ \eta $ denotes the learning rate, $ \lambda $ is the regularization coefficient, and $ \mathbf{W}^{(l, n)} $ represents the weight matrix of the $ l $-th layer at the $ n $-th iteration. The term $ \nabla_{\mathbf{W}^{(l)}} L_{\text{CE}} $ is the gradient of the cross-entropy loss with respect to the weights of the $ l $-th layer. $ - \eta \lambda \mathbf{W}^{(l, n)} $  serves as a regularization term, constraining the 2-norm of the weight matrix at each layer within a bounded range.

Therefore, assume there exist constants $ C_1^{(l)} $ and $ C_2^{(l)} $ such that:
\begin{equation}
    \vspace{+0.1cm}
    C_1^{(l)} \leq \| \mathbf{W}^{(l)} \|_2 \leq C_2^{(l)} \label{eq6}
    \vspace{+0.1cm}
\end{equation}
given that $S^{(l-1,t)}\in \left \{0,1  \right \}$ and $I^{(l,t)} = W^{(l)} S^{(l-1,t)}+b^{(l)}$, we have:
\begin{equation}
    \vspace{+0.1cm}
    {I}^{(l,t)}_{min} \leq {I}^{(l,t)} \leq{I}^{(l,t)}_{max} \label{eq7}
    \vspace{+0.1cm}
\end{equation}

When the neuron receives only inhibitory connections, the input current tends towards its minimum value; conversely, when it receives only excitatory connections, the input current tends towards its maximum value. Therefore:
\begin{equation}
\vspace{+0.1cm}
    \begin{cases}
        {I}^{(l)}_{max} = \sum_{i}^{C}\sum_{j}^{H}\sum_{k}^{W}  relu(\mathbf{W}_{ijk}^{(l)})+\mathbf{b}^{(l)}\\
        {I}^{(l)}_{min} =-\sum_{i}^{C}\sum_{j}^{H}\sum_{k}^{W} relu(-\mathbf{W}^{(l)}_{ijk})+\mathbf{b}^{(l)}
    \end{cases}
    \label{eq8}
\vspace{+0.1cm}
\end{equation}

From Equation (\ref{eq2}), it follows that the membrane potential increases monotonically in the absence of inhibitory inputs. The minimum membrane potential, $ U_{\min} $, is maintained throughout the time period $ T $, during which the neuron receives only inhibitory currents and no excitatory inputs. This paper adopts a hard reset mechanism, ensuring that the membrane potential does not exceed the threshold voltage at any given moment. 
\begin{equation}
\vspace{+0.1cm}
    \begin{cases}
        U^{(l)}_{max} < U_{th} \\
        U^{(l)}_{\min} = U_{\min}^{(l,T)} =  \frac{\left( 1 - \frac{1}{\tau_m^{T-1}} \right) {I}^{(l)}_{\min}}{1 - \frac{1}{\tau_m}}<  \frac{\tau_m }{\tau_m - 1}{I}^{(l)}_{\min}  \\
        % \hspace{2.16cm} <  \frac{\tau_m }{\tau_m - 1}{I}^{(l)}_{\min}
    \end{cases}
    \label{eq9}
\vspace{+0.1cm}
\end{equation}

Thus, based on the LIF dynamics described in Equation (\ref{eq1}), we can derive the maximum and minimum values of the membrane potential derivative $ U'(t) $:
\begin{equation}
    \begin{cases}
        U'^{(l)}_{max} = - \frac{1}{\tau_m}(U^{(l)}_{max} -U_{\text{rest}}) +  {I}^{(l)}_{min}\\
        U'^{(l)}_{min} = - \frac{1}{\tau_m}(U^{(l)}_{min} -U_{\text{rest}}) +  {I}^{(l)}_{max}
    \end{cases}
    \label{eq10}
\end{equation}

Since both the input current and membrane potential are bounded functions, by substituting into Equation (\ref{eq10}), we obtain $ U'{\text{max}} $ and $ U'{\text{min}} $, indicating that the membrane potential derivative is also bounded.

\subsection{ Historical Dynamic Refractory Period}

The refractory period is commonly used to mitigate the problem of neuronal over-activation. The degree of neuronal over-activation needs to be determined by comprehensively considering both the state of the neuron's membrane potential and the history of spike firing rates. The larger the derivative of the neuronal membrane potential, the more dramatic the change in membrane potential, indicating a higher degree of neuronal over-activation. Therefore, we can dynamically adjust the length of the refractory period based on the membrane potential derivative. The longer the refractory period, the stronger its role in inhibiting the accumulation state of the neuron's membrane potential. For neurons that still exhibit strong over-activation despite refractory period inhibition, it is necessary to further slow down the recovery rate of the refractory period decay to zero, thereby extending their inhibition time.

Meanwhile, the historical refractory period can implicitly represent the history of the neuron's spike firing rates. The higher the historical spike firing rate, the greater the accumulated value of the historical refractory period. Therefore, this paper proposes the Historical Dynamic Refractory Period (HDRP), denoted as $\tau_{ref}^{(l,t)}$, to ensure the continuity and effectiveness of the inhibitory mechanism over multiple time steps, thereby significantly reducing the negative impact of noise on neuronal activity. The definition of HDRP is as follows:
\begin{equation}
    \begin{split}
        \tau_{ref}^{(l,t)}   &=   k({U'}^{(l,t)})(\tau_{ref}^{(l,t-1)} +{S}^{(l,t)}\tau_{ref_{L}})\\
        &\quad+{S}^{(l,t)}\tau_{ref_{0}}
    \end{split}
    \label{eq11}
    \vspace{+0.2cm}
\end{equation}

where $\tau_{ref} \in (\tau_{ref_{0}}, \tau_{ref_{max}})$, $\tau_{ref_{0}}$ is the minimum refractory period, and $\tau_{ref_{L}}$ is the length of the refractory period range, representing the adjustable range of the refractory period. $ k(U'^{(l,t)} ) $is the membrane potential derivative normalization function.

In Section \ref{subsec:4-1bounds}, we derived the upper and lower bounds of the membrane potential derivative, showing that the value range of the membrane potential derivative is entirely determined by the synaptic connections pre-established between neurons, independent of the input signal. Therefore, we employ the Min-Max normalization method to normalize the membrane potential derivative, which not only simplifies the computation and ensures consistent scaling but also avoids issues such as data distortion and gradient vanishing caused by normalization methods like the Sigmoid function. The normalization function $ k(U'^{(l,t)}) $ is defined as follows:
\begin{equation}
    k({U'}^{(l,t)})=\frac{U'^{(l,t)}-U'^{(l)}_{min}}{U'^{(l)}_{max}-U'^{(l)}_{min}} 
    \label{eq12}
    \vspace{+0.2cm}
\end{equation}

\subsection{HDRP-SNN Model}

In HDRP, the longer the refractory period, the higher the degree of neuronal over-activation, making it more difficult for the neuron to fire spikes. The function used to describe this inhibitory effect is called the refractory kernel. To simulate the inhibitory effect of the relative refractory period, we designed the refractory kernel as a threshold-dependent decreasing function, meaning that as the refractory period shortens, the inhibition effect gradually weakens. Due to the modulation by HDRP, the refractory kernel can flexibly regulate the neuron's firing state. Therefore, the refractory kernel for HDRP is defined as follows:
\begin{equation}
    g(\tau_{\text{ref}}) = U_{th}*tanh(A\tau_{{ref}})
    \label{eq13}
\end{equation}
where $A>0$, and $A$ is a learnable parameter. $U_{th}$ represents the neuron's threshold voltage.

According to the refractory kernel, we derive a new synaptic current expression:

\begin{equation}
    I^{(l,t)} =  W^{(l)}  S^{(l-1,t)} + b -g(\tau_{ref}^{(l,t-1)})
    \label{eq14}
\end{equation}

For a clearer computational representation, LIF with HDRP (HDRP-LIF) can be shown iteratively expressed in SNNs as:
\begin{equation}
    \begin{cases}
        \mathbf{I}^{(l,t)}=\mathbf{W}^{(l)}\mathbf{S}^{(l-1,t)}+ \mathbf{b} -g(\boldsymbol{\tau}_{ref}^{(l,t-1)})\\
        \mathbf{U'}^{(l,t)} =  - \frac{1}{\tau_m}(\mathbf{U}^{(l,t-1)} -U_{\text{rest}}) +  \mathbf{I}^{(l,t)}\\
        \mathbf{H}^{(l,t)} = \mathbf{U}^{(l,t-1)}+\mathbf{U'}^{(l,t)}\\
        \mathbf{S}^{(l,t)} = \Theta(\mathbf{H}^{(l,t)} - U_{th}) \\
        \mathbf{U}^{(l,t)} =  \mathbf{H}^{(l,t)} \cdot (1 - \mathbf{S}^{(l,t)}) + U_{\text{rest}} \mathbf{S}^{(l,t)}\\
        % \mathbf{K}^{(l,t)} = k(\mathbf{U'}^{(l,t)})\\
        \boldsymbol{\tau}_{ref}^{(l,t)} =   k(\mathbf{U'}^{(l,t)})\boldsymbol{\tau}_{ref}^{(l,t-1)}\\ \hspace{1.16cm}+\mathbf{S}^{(l,t)}(k(\mathbf{U'}^{(l,t)})\tau^{(l)}_{ref_{L}}+\tau^{(l)}_{ref_{0}})\\
    \end{cases}
    \label{eq15}
\end{equation}

based on Equation (\ref{eq15}), HDRP-LIF model can be directly interchanged with LIF model in traditional SNNs without destroying the binary nature of SNNs, ensuring the locally independent nonlinear computation of neurons. The architecture HDRP-SNN is shown in Fig. \ref{fig:1}.

\subsection{Training HDRP-SNN}

The expressions and update process of the HDRP-SNN model parameters $W$ and $A$ in Spatio-temporal backpropagation (STBP) \cite{wu2018spatio} training are derived in this paper, and the specific formulas are as follows: 
\begin{equation}
    \begin{cases}
        \delta \mathbf{S}^{(l,t)} = 
        \begin{cases} 
        \delta \mathbf{S}^{(L,t)}, &\!\!\!\!\! \text{if } l = L \\
        \delta \mathbf{H}^{(l+1,t)} \mathbf{W}^{(l+1)}+\\
        \delta \mathbf{U}^{(l,t)}(U_{rest}-\mathbf{H}^{(l,t)}), &\!\!\!\!\! \text{if } l \in (1,L-1)\\
        \end{cases} \\
        
        \delta \boldsymbol{\tau}_{{ref}}^{(l,t+1)} = 
            \delta \boldsymbol{\tau}_{{ref}}^{(l,t+2)} k(\mathbf{U}'^{(l,t+2)}) 
            + \delta \mathbf{H}^{(l,t+2)} \\
            \hspace{1.7cm}(-U_{{th}} \mathbf{A}^{(l)} 
            (1 - tanh^2(\mathbf{A}^{(l)} \boldsymbol{\tau}_{{ref}}^{(l,t+1)})))\\
            
        \delta \mathbf{U'}^{(l,t+1)} = \delta \mathbf{H}^{(l,t+1)} + \delta \boldsymbol{\tau}_{{ref}}^{(l,t+1)}\\
        
        \delta \mathbf{U}^{(l,t)} = \delta \mathbf{H}^{(l,t+1)} - \frac{\delta \mathbf{U'}^{(l,t+1)}}{\tau_m}\\
        
        \delta \mathbf{H}^{(l,t)} = \delta \mathbf{S}^{(l,t)} \sigma' + \delta \mathbf{U}^{(l,t)} (1 - \mathbf{S}^{(l,t)} \\
        \hspace{1.4cm}+ \sigma'(U_{\text{rest}} - \mathbf{H}^{(l,t)}))\\
        
        \delta \mathbf{W}^{(l)} = \mathbf{S}^{(l-1,t)} \delta \mathbf{H}^{(l,t)}\\

        \delta \mathbf{A}^{(l)} = U_{{th}} \tau_{{ref}}^{(l,t-1)} 
        (tanh^2(\mathbf{A}^{(l)} \tau_{{ref}}^{(l,t-1)})-1) \delta \mathbf{H}^{(l,t)}\\
        
        \delta \mathbf{b}^{(l)} = \delta \mathbf{H}^{(l,t)}\\
        
    \end{cases}
    \label{eq16}
\end{equation}

where $\delta$ is the gradient of variables with respect to the loss function $L_\text{total}$, and $\sigma'$ is the gradient surrogate function for the non-differentiable spike activation in Equation (\ref{eq17}):
\begin{equation}
\vspace{+0.1cm}
    \sigma'=\frac{\partial \mathbf{S}^{(l,t)}}{\partial \mathbf{H}^{(l,t)}}=\frac{1}{\gamma^2} \max \left( 0, \gamma - \left| \mathbf{H}^{(l,t)} - U_{th} \right| \right)
    \label{eq17}
\vspace{+0.1cm}
\end{equation}

\subsection{Operations for HDRP-SNN}
\label{subsec:energy_efficiency}

For ANNs, computational cost mainly comes from multiply-accumulate (MAC) operations in matrix multiplications. In contrast, SNNs rely on an addressing mechanism, where computation depends only on address updates. This allows SNNs to perform accumulate (AC) operations instead of MACs in convolutional and fully connected layers. Therefore, all spike-related multiplication operations in SNNs can be treated as AC operations.

In SNNs, MAC operations primarily stem from temporal neuron state updates and encoding layers, while all other operations are treated as AC operations. Consequently, neuron MAC operations are determined solely by network synaptic connections. AC operations are influenced by both synaptic connections (fan-out $f_i^l$ representing the number of outgoing connections) and neuron spikes $s_i^l[t]$. The definitions of ACs and MACs are as follows:
\begin{equation}
    \begin{cases}
        \text{ACs} = \sum_{t=1}^{T} \sum_{l=1}^{L-1} \sum_{i=1}^{N_l} f_i^l s_i^l[t]  \\
        \text{MACs} = \sum_{t=1}^{T}\sum_{l=1}^{L-1} \sum_{i=1}^{N_l} f_i^l 
    \end{cases}
    \label{eq23}
\end{equation}

It is assumed that a 32-bit floating-point implementation is used under 45nm technology, with the energy consumption of various operations defined as $ E_{MAC} = 4.6 \, pJ $ and $ E_{AC} = 0.9 \, pJ $. The total energy consumption for synaptic operations in SNNs and ANNs is as follows:

\begin{equation}
    E_{total} = MAC*E_{MAC} +  AC*E_{AC}
    \label{eq24}
\end{equation}

% 精度
\begin{table}[b]
    \centering
    \small
    \begin{threeparttable}

    \setlength{\tabcolsep}{0.4pt}
    \begin{tabular}{@{}c@{}c@{}c@{}ccc}
        \toprule
        \textbf{Dataset} & \textbf{Method} & \textbf{Architecture} & \textbf{Neuron} & \textbf{T} & \textbf{Acc(\%)} \\
        \midrule
        
        \multirow{9}{*}{\rotatebox{90}{\textbf{CIFAR10}\tnote{1}}} 
            & ANN\tnote{*} & ResNet-18 & ReLU & 1 & 96.49 \\ 
            \cmidrule{2-6}
            
            & PSN \cite{fang2024parallel} & Modified PLIF Net & PSN & 4 & 95.32 \\ 
            & TAB \cite{jiangtab} & ResNet-19 & LIF & 6 & 94.81 \\
            & LOCALZO \cite{mukhoty2023direct} & ResNet-19 & LIF & 6 & 95.56 \\
            & DSR \cite{meng2022training} & PreAct-ResNet-18 & LIF & 20 & 95.40 \\
            & LM-H \cite{hao2023progressive} & ResNet-19 & LM-H & 4 & 96.36 \\
            \cmidrule{2-6}
            
            & \multirow{3}{*}{Ours} & ResNet-18 & HDRP-LIF & 2 & \textbf{96.14} \\ 
            & ~ & ResNet-18 & HDRP-LIF & 4 & \textbf{96.56} \\
            & ~  & ResNet-18 & HDRP-LIF & 6 & \textbf{96.70} \\
        \midrule
        
        \multirow{8}{*}{\rotatebox{90}{\textbf{CIFAR100}\tnote{1}}} 
            & ANN\tnote{*} & ResNet-18 & ReLU & 1 & 80.53 \\
            \cmidrule{2-6}
            
            & TAB \cite{jiangtab} & ResNet-19 & LIF & 6 & 76.82 \\
            & LOCALZO \cite{mukhoty2023direct} & ResNet-19 & LIF & 6 & 77.25 \\
            & DSR \cite{meng2022training} & PreAct-ResNet-18 & LIF & 20 & 78.50 \\
            & LM-H \cite{hao2023progressive} & ResNet-19 & LM-H & 4 & 80.31 \\
            \cmidrule{2-6}
            
            & \multirow{3}{*}{Ours} & ResNet-18 & HDRP-LIF & 2 & \textbf{79.61} \\ 
            &  & ResNet-18 & HDRP-LIF & 4 & \textbf{80.34} \\
            & & ResNet-18 & HDRP-LIF & 6 & \textbf{80.91} \\
        \midrule
        
        \multirow{6}{*}{\rotatebox{90}{\textbf{ImageNet-1k}\tnote{1}}} 
            & ANN\tnote{*} &  ResNet-34 & ReLU & 1 & 73.30 \\  
            \cmidrule{2-6}
            
            & PSN \cite{fang2024parallel} & SEW ResNet-34 & PSN & 4 & \textbf{70.54} \\ 
            & TAB \cite{jiangtab} & ResNet-34 & LIF & 4 & 67.78 \\
            & DSR \cite{meng2022training} & PreAct-ResNet-18 & IF & 50 & 67.74 \\
            & MS-ResNet \cite{hu2024advancing} & ResNet-34 & LIF & 6 & 69.42 \\
            & LM-H \cite{hao2023progressive} & ResNet-34 & LM-H & 4 & 69.73 \\
            \cmidrule{2-6}
            
            & Ours & ResNet-34 & HDRP-LIF & 6 & 70.45 \\ 
        \midrule 
        
        \multirow{6}{*}{\rotatebox{90}{\textbf{CIFAR10-DVS}\tnote{1}}} 
            & PSN \cite{fang2024parallel} & VGG & sliding PSN & 10 & 85.90 \\
            & LOCALZO \cite{mukhoty2023direct} & VGGSNN & LIF & 10 & 81.87 \\
            & TAB \cite{jiangtab} & 7-layer CNN & LIF & 4 & 76.7 \\
            & DSR \cite{meng2022training} & VGG-11 & LIF & 20 & 77.27 \\
            & LM-H \cite{hao2023progressive} & ResNet-19 & LM-H & 10 & 79.10 \\
            \cmidrule{2-6}
            
            & Ours & VGGSNN & HDRP-LIF & 10 & \textbf{87.00}  \\ 
        \bottomrule
    \end{tabular}
    \begin{tablenotes}
        \item[*] The self-implemented ANN shares identical structures and hyper-parameters with SNN.
        \item[1] Applied data augmentation method CutMix\cite{yun2019cutmix}.
    \end{tablenotes}
    \caption{Performance of HDRP-SNN in static and neuromorphic dataset.}
    \label{tab1}
    \end{threeparttable}
    \vspace{-0.4cm}
\end{table}

\section{Experiment}

To validate the HDRP-SNNs, we conduct experiments on three static datasets (CIFAR10, CIFAR100, and ImageNet-1k) and one neuromorphic dataset (CIFAR10-DVS) for classification tasks to assess model performance and energy efficiency, along with an analysis of noise resilience under varying levels of Gaussian noise. An ablation study further examines the impact of different refractory period settings.

\subsection{Performance Analysis}

We integrated HDRP-LIF neurons into various network architectures and evaluated their performance on both static and neuromorphic datasets. As shown in Table \ref{tab1}, HDRP-SNN consistently outperforms current state-of-the-art methods. On CIFAR10 and CIFAR100, our approach achieves competitive accuracy even at $T=2$ and surpasses all other SNN methods at $T=4$, reaching 96.56\% and 80.34\% accuracy, respectively. At $T=6$, HDRP-SNN exceeds the corresponding ANN benchmark within the same architecture by achieving 96.70\% on CIFAR10 and 80.91\% on CIFAR100.

On the more challenging ImageNet-1k dataset, HDRP-SNN attains 70.45\% accuracy at $T=6$, which is comparable to the best-reported SOTA performance (70.54\%) and offers improvements of 2.67\% and 1.03\% over the TAB and MS-ResNet baselines, respectively, when using the ResNet-34 architecture. Furthermore, on the neuromorphic CIFAR10-DVS dataset, our VGGSNN-based model sets a new record by achieving 87.00\% accuracy, significantly outperforming the closest competitor (PSN), which reaches only 85.90\%.

% 能耗
\begin{table*}[b]
    \centering
    \small
    \setlength{\tabcolsep}{8pt}

    \begin{tabular}{ccccccc}
        \toprule      

        \textbf{Dataset} & \textbf{Neuron} & \textbf{T} & \textbf{Parameters (M)} & \textbf{ACs (M)} & \textbf{MACs (M)} & \textbf{SOP Energy (µJ)}\\
        
        \midrule
        \multirow{3}{*}{\shortstack{\textbf{CIFAR10}\\(ResNet18)}}
            & ReLU              & 1 & 12.50 & 0.00                  & 2191.20   & 10079.52\\ 
            & LIF               & 6 & 12.50 & 128.27                & \textbf{8.25}      & \textbf{153.43}\\
            & HDRP-LIF (Ours)   & 6 & 12.50 & \textbf{105.85}       & 16.51     & 171.23\\ 
        \midrule
        
        \multirow{3}{*}{\shortstack{\textbf{CIFAR100}\\(ResNet18)}}
            & ReLU              & 1 & 12.50 & 0.00                  & 2191.20   & 10079.52\\
            & LIF               & 6 & 12.50 & 142.10                & \textbf{8.25}      & \textbf{165.87}\\
            & HDRP-LIF (Ours)   & 6 & 12.50 & \textbf{120.68}       & 16.51     & 184.58\\ 
        \midrule   
        
        \multirow{3}{*}{\shortstack{\textbf{ImageNet}\\(ResNet34)}}
            & ReLU              & 1 & 21.80 & 0.00                  & 3644      & 16762.40\\ 
            & LIF               & 6 & 21.80 & 323.38                & \textbf{21.38}      & \textbf{389.40}\\
            & HDRP-LIF (Ours)   & 6 & 21.80 & \textbf{267.05}       & 42.76     & 437.05\\ 
        \midrule
        
        \multirow{2}{*}{\shortstack{\textbf{CIFAR10-DVS}\\(VGG11)}}  
            & LIF               & 10 & 9.27 & 255.92                & \textbf{3.41}     & \textbf{246.01}\\ 
            & HDRP-LIF (Ours)   & 10 & 9.27 & \textbf{245.86}       & 6.82     & 252.65\\ 
        \bottomrule
    \end{tabular}
    \caption{Energy performance of HDRP-SNN in static and neuromorphic dataset.}
    \label{tab2}
    \vspace{-0.2cm}
\end{table*}

Although introducing the historical dynamic refractory period into traditional LIF neurons increases model complexity to a certain extent, the mechanism's capability to suppress neuronal over-activation does not degrade performance. Instead, it achieves superior or comparable accuracy across multiple datasets. Specifically, compared to conventional SNN models without refractory mechanisms, HDRP-LIF maintains higher or comparable accuracy under the same or fewer time steps. This indicates that incorporating HDRP facilitates more precise regulation of neuronal firing behavior, thereby enhancing the network’s ability for feature learning and representation, ultimately leading to significant overall performance improvements.

\subsection{Energy Efficiency Analysis}

To evaluate the energy efficiency of the HDRP mechanism, we adopted the energy consumption evaluation method described in Section 4.5 to systematically quantify the parameter count, accumulation (AC) operations, multiply-accumulate (MAC) operations, and SOP energy consumption of ANN, SNN, and HDRP-SNN across four datasets (CIFAR10, CIFAR100, ImageNet-1k, and CIFAR10-DVS) in Table \ref{tab2}. In SNNs, MAC operations primarily originate from the state updates of temporal neurons and the encoding layer, with neuron-specific MAC operations determined solely by the number of synaptic connections in the network. In contrast, AC operations are influenced by both the synaptic connectivity and the firing rate of neurons; for a given network architecture, a lower number of AC operations indicates a reduced firing rate.

Although the model formulation of HDRP-SNN suggests an increased computational complexity in its neuron model compared to conventional SNNs, the statistical results demonstrate that the parameter count of HDRP-SNN is essentially comparable to that of ANN and SNN, indicating that the incorporation of HDRP does not significantly increase the overall number of network parameters. Furthermore, the overall firing rate in HDRP-SNN is modulated by multiple factors including pre-synaptic inputs, training weights, neuron parameters, and the refractory kernel. The inherent pulse-suppression mechanism in HDRP-SNN prolongs the membrane potential accumulation process, thereby reducing the overall firing frequency, which results in a lower AC operation count in HDRP-SNN compared to LIF-SNN across the evaluated datasets. In the SOP energy consumption analysis, despite the increased computational complexity of HDRP-SNN, the reduction in AC operations leads to an energy consumption that is only marginally higher than that of LIF-SNN, while still maintaining a significant energy advantage over ANN.

In summary, the integration of the HDRP mechanism not only enhances the representational capacity of LIF-SNN but also preserves its inherent low-energy consumption advantage.

\subsection{Noise Resistance Analysis}

\begin{figure}[h]
  \centering
   \includegraphics[width=1\linewidth]{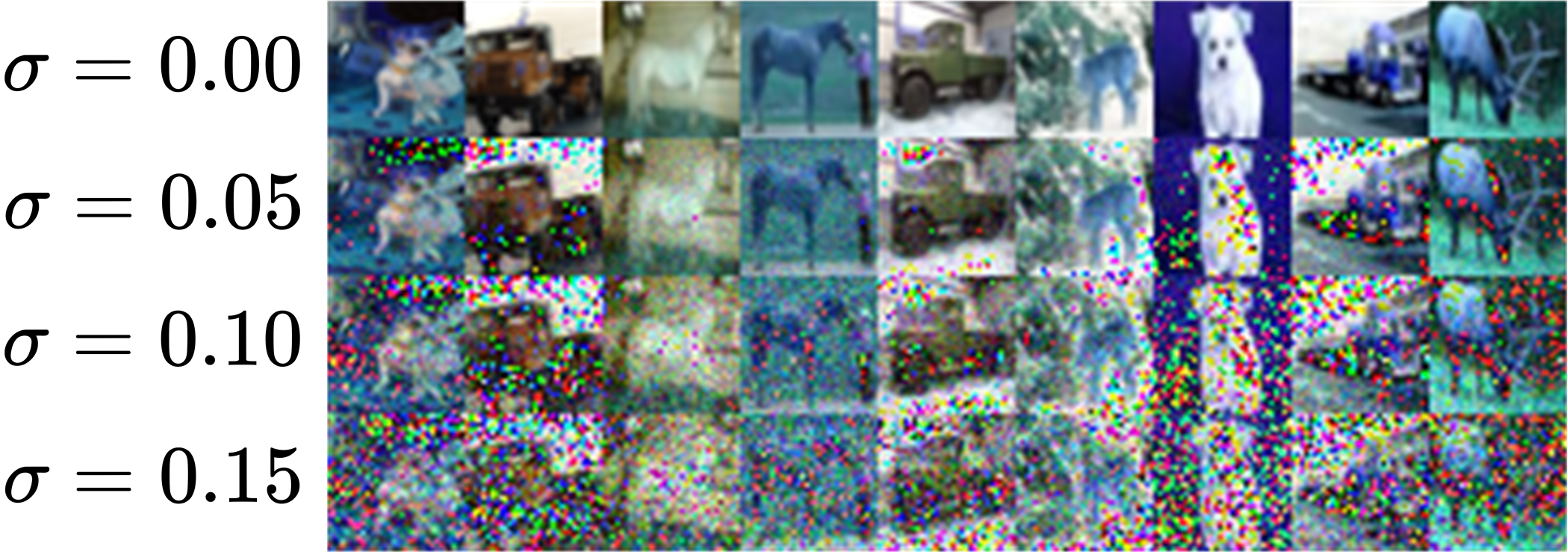}
   \caption{Examples of CIFAR10 Images Under Different Levels of Gaussian Noise.}
   \label{fig:2}
   \vspace{-0.2cm}
\end{figure}

Neuronal over-activation can lead to overall network chaos and signal interference, thereby compromising the accuracy and stability of information transmission. The refractory period mechanism effectively mitigates the issue of over-activation, and it is worth investigating whether this mechanism can also enhance a model’s robustness to noisy inputs. To validate the noise robustness of the models, we conducted experiments using identical network architectures (ResNet18 and VGGSNN) on both a static dataset (CIFAR10) (see Fig.\ref{fig:2}) and a neuromorphic dataset (CIFAR10-DVS), with various levels of added Gaussian noise, evaluating the classification accuracy of ANN, SNN, and HDRP-SNN.

\begin{figure*}[b]
    \centering
    \begin{subfigure}[t]{0.5\textwidth}
        \centering
        \includegraphics[width=0.96\textwidth]{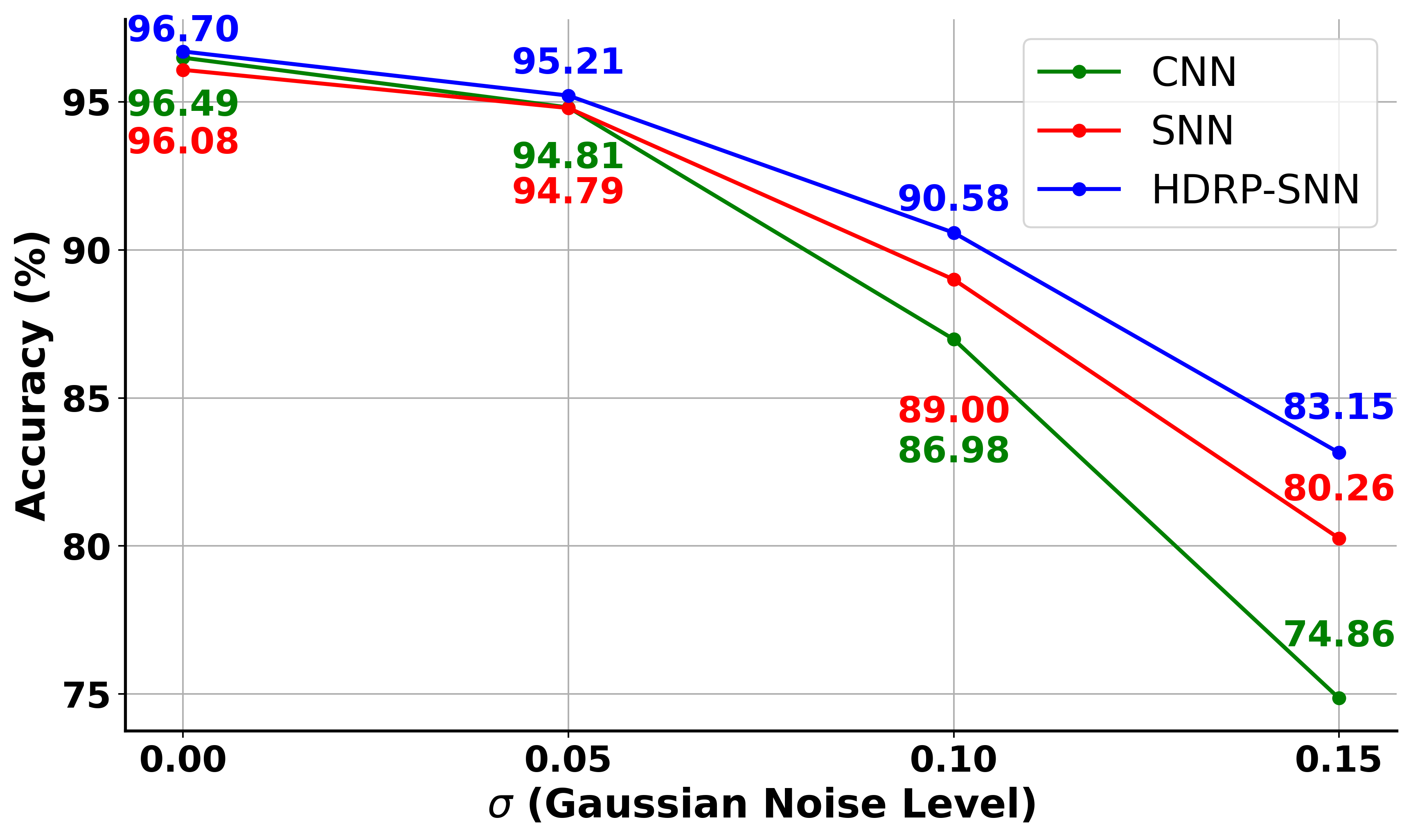}
        \caption{CIFAR10}
        \label{fig:3(a)}
    \end{subfigure}%
    ~
    \begin{subfigure}[t]{0.5\textwidth}
        \centering
        \includegraphics[width=0.96\textwidth]{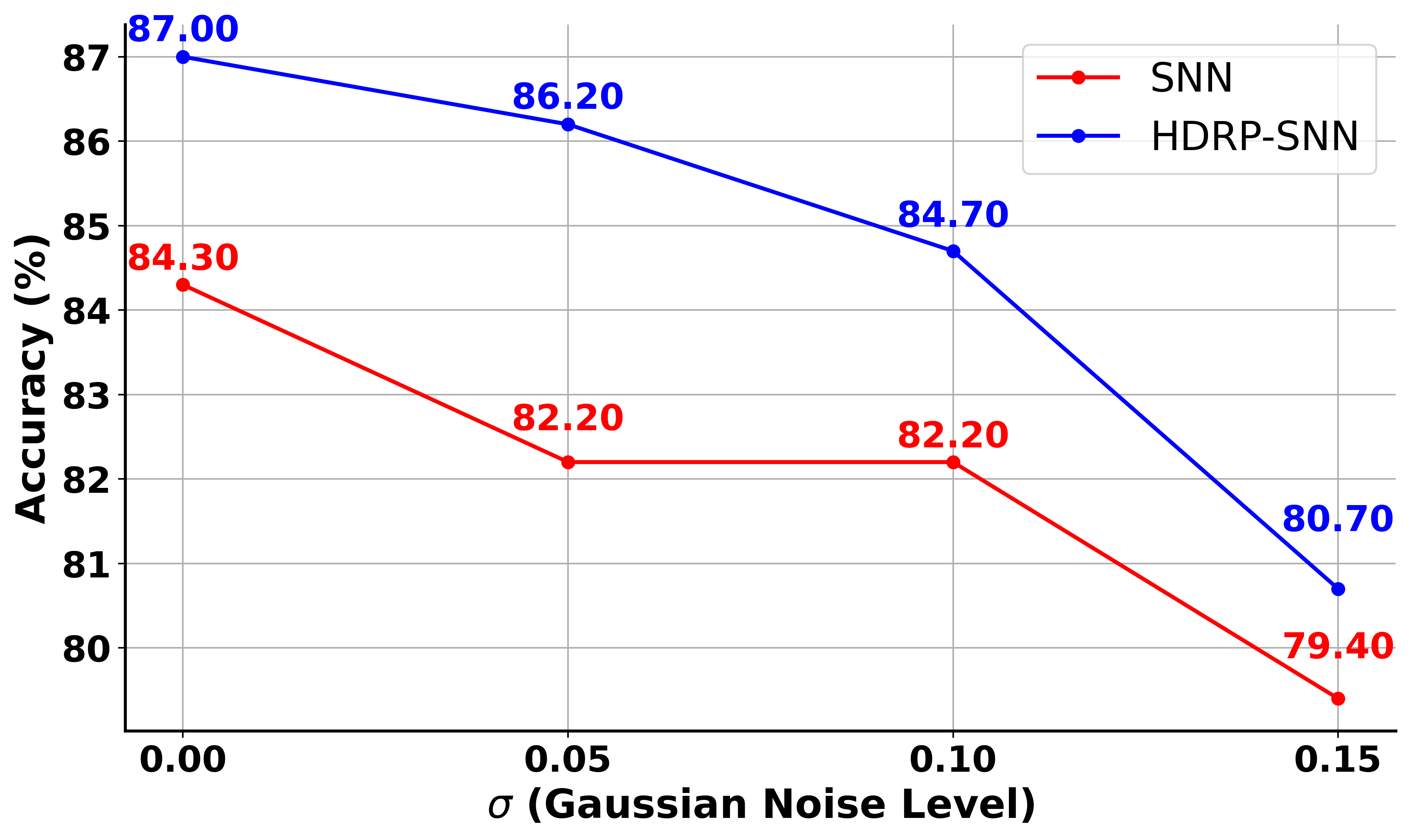}
        \caption{CIFAR10-DVS}
        \label{fig:3(b)}
    \end{subfigure}

   \caption{Accuracy Comparison of Different Models Under Varying Gaussian Noise Levels.}
   \label{fig:3}
   \vspace{-0.4cm}
\end{figure*}

In the noise robustness experiments on the static CIFAR10 dataset (see Fig.\ref{fig:3(a)}), HDRP-SNN consistently exhibited superior noise resilience across all Gaussian noise levels ($\sigma$ values), maintaining high classification accuracy throughout. Notably, at the highest noise level, HDRP-SNN achieved accuracy improvements of 2.89\% and 8.29\% over SNN and ANN, respectively. The trend in the accuracy curves indicates that ANN is most sensitive to noise, likely due to the absence of the spike-based filtering mechanism inherent in SNNs. Although standard SNNs demonstrate better robustness than CNNs, their noise resistance remains significantly inferior to that of HDRP-SNN.

In the noise robustness experiments on the more complex neuromorphic CIFAR10-DVS dataset (see Fig.\ref{fig:3(b)}), despite observable performance fluctuations across different noise levels, HDRP-SNN still exhibited stronger noise robustness compared to standard SNNs, achieving up to a 4.2\% improvement in accuracy.
In summary, the experimental results demonstrate that the incorporation of the HDRP mechanism enhances the model’s ability to effectively filter out noisy inputs, significantly improving its noise resistance and performance stability while further amplifying the inherent noise robustness conferred by spike-based processing in SNNs.

\begin{table*}[h!]
    \begin{center}
    \centering
    \small
    \setlength{\tabcolsep}{8pt}
    \begin{threeparttable}
        \begin{tabular}{ccccccccc}
            \toprule 
            \textbf{Model} & \textbf{LIF} & \multicolumn{2}{c}{\textbf{LIF}} & \textbf{no His}\tnote{1} & \multicolumn{3}{c}{\textbf{His with Fixed Decay Factor}\tnote{2}} & \textbf{Ours} \\
                        
            \textbf{Refractory Period Type} & - & \multicolumn{2}{c}{absolute} & relative & \multicolumn{3}{c}{relative} & relative \\

            \cmidrule{3-4}
            \textbf{Refractory Period Length} & - & 1 & 2 & adaptive, $\le 6$ & \multicolumn{3}{c}{adaptive, $\le 6$} & adaptive, $\le 6$ \\

            \textbf{Decay Type} & - & \multicolumn{2}{c}{linear} & exponential & \multicolumn{3}{c}{exponential} & exponential \\   

            \cmidrule{3-4}
            \cmidrule{6-8}
            \textbf{Decay Factor} & - & 1 & 1 & adaptive \tnote{3} & 0.1 & 0.5 & 0.9 & adaptive \tnote{3}\\
            \midrule
            
            \textbf{CIFAR-10 (\%)} & 96.08 & 94.40 & 93.90& 96.39 & 95.62 & 96.34 & 93.02 & \textbf{96.70} \\
            
            \textbf{CIFAR10-DVS (\%)} & 84.31 & 84.40& 80.40& 85.82 & 85.84 & 85.94 & 75.00 & \textbf{87.00} \\
            \bottomrule
        \end{tabular}
        \begin{tablenotes}
            \item[1] No historical refractory period.
            \item[2] Historical refractory period with the fixed decay factor.
            \item[3] Adaptive dynamic decay based on the membrane potential derivative.
        \end{tablenotes}
    \end{threeparttable}
    \end{center}
    \vspace{-0.4cm}
    \caption{Ablation Study of HDRP-LIF}
    \label{tab3}
    \vspace{-0.1cm}
\end{table*}

\subsection{Ablation Study}

To validate the effectiveness of the two core components in the HDRP-LIF model—adaptive dynamic decay and dynamic historical refractory period—we conducted a series of ablation studies on the CIFAR-10 and CIFAR10-DVS datasets. Table \ref{tab3} presents the classification accuracies achieved under various configurations.

As a baseline, we employed a conventional LIF model without any decay or refractory mechanisms, which yielded accuracies of 96.08\% on CIFAR-10 and 84.31\% on CIFAR10-DVS. We then introduced an absolute refractory period (with durations of 1 and 2 time steps) combined with a linear decay mechanism. However, this modification degraded performance, indicating that an absolute refractory period may overly constrain the neuronal dynamic responses.

Subsequently, we adopted a relative refractory period (adaptive, with a maximum length of 6 time steps) alongside an exponential decay mechanism. Removing the historical refractory period altogether resulted in improved accuracies of 96.39\% and 85.82\% on CIFAR-10 and CIFAR10-DVS, respectively. Furthermore, when incorporating a historical refractory period with a fixed decay factor (0.1, 0.5, and 0.9)\cite{higuchi2024balanced}, only the configuration with a decay factor of 0.5 achieved competitive performance, while other fixed values led to inferior outcomes. These results highlight the sensitivity of fixed decay strategies to parameter selection.

In contrast, the proposed HDRP-LIF model synergistically integrates adaptive dynamic decay with a dynamic historical refractory period. By leveraging the membrane potential derivative to adjust the decay factor adaptively and by dynamically modulating the historical refractory period, HDRP-LIF achieved the highest accuracies of 96.70\% on CIFAR-10 and 87.00\% on CIFAR10-DVS. This confirms that the combined approach effectively enhances the temporal response of neurons and improves overall model performance on visual recognition tasks.

The ablation experiments demonstrate that the integration of adaptive dynamic decay and dynamic historical refractory period in HDRP-LIF not only overcomes the limitations of fixed decay strategies but also optimally modulates neuronal temporal dynamics, providing robust support for its efficacy in visual recognition applications.

As shown in Table \ref{tab3}, HDRP-SNN achieved the highest accuracy of 96.70\%, using the LIF model without a refractory period as the benchmark for this ablation study. Adding dynamic decay improved accuracy by 0.31\% over the benchmark, demonstrating the benefits of dynamic decay based on membrane potential derivatives. However, omitting the historical refractory period resulted in lower accuracy compared to HDRP-SNN, emphasizing the importance of incorporating historical context in regulating neuronal activity. Models with fixed decay factors showed significant fluctuations in performance, with an initial rise followed by a decline, indicating that the choice of decay factor strongly affects network stability and overall performance.

Ablation experiments demonstrated that the HDRP mechanism effectively adapts to network states by leveraging both current neuron state and historical refractory period, exhibiting strong adaptability in low-latency SNNs. In contrast, fixed refractory period adjustment mechanisms struggle to regulate these networks effectively. This highlights the crucial role of the HDRP mechanism in enhancing the performance of low-latency SNNs.

\section{Conclusion}

This paper introduces HDRP-SNN, a low-latency spiking neural network that incorporates a historical dynamic refractory period mechanism. This mechanism enables neurons to self-regulate locally by dynamically adjusting their excitability based on both their current state and their refractory history, thereby facilitating precise and rapid modulation within fewer time steps. Extensive experiments demonstrate that HDRP-SNN achieves state-of-the-art accuracy in image classification tasks while exhibiting superior noise robustness compared to both traditional SNNs and ANNs with equivalent architectures. By retaining the inherent computational benefits of SNNs, HDRP-SNN seamlessly integrates into existing LIF-SNN architectures. Moreover, it effectively reduces spiking firing rate, thereby enhancing both expressiveness and energy efficiency, especially in large-scale networks.

{
    \small
    \bibliographystyle{ieeenat_fullname}
    \bibliography{main}

\begin{thebibliography}{45}
\providecommand{\natexlab}[1]{#1}
\providecommand{\url}[1]{\texttt{#1}}
\expandafter\ifx\csname urlstyle\endcsname\relax
  \providecommand{\doi}[1]{doi: #1}\else
  \providecommand{\doi}{doi: \begingroup \urlstyle{rm}\Url}\fi

\bibitem[Angelidis et~al.(2021)Angelidis, Buchholz, Arreguit, Roug{\'e}, Stewart, von Arnim, Knoll, and Ijspeert]{angelidis2021spiking}
Emmanouil Angelidis, Emanuel Buchholz, Jonathan Arreguit, Alexis Roug{\'e}, Terrence Stewart, Axel von Arnim, Alois Knoll, and Auke Ijspeert.
\newblock A spiking central pattern generator for the control of a simulated lamprey robot running on spinnaker and loihi neuromorphic boards.
\newblock \emph{Neuromorphic Computing and Engineering}, 1\penalty0 (1):\penalty0 014005, 2021.

\bibitem[Avissar et~al.(2013)Avissar, Wittig, Saunders, and Parsons]{avissar2013refractoriness}
Michael Avissar, John~H Wittig, James~C Saunders, and Thomas~D Parsons.
\newblock Refractoriness enhances temporal coding by auditory nerve fibers.
\newblock \emph{Journal of Neuroscience}, 33\penalty0 (18):\penalty0 7681--7690, 2013.

\bibitem[Bellec et~al.(2020)Bellec, Scherr, Subramoney, Hajek, Salaj, Legenstein, and Maass]{bellec2020solution}
Guillaume Bellec, Franz Scherr, Anand Subramoney, Elias Hajek, Darjan Salaj, Robert Legenstein, and Wolfgang Maass.
\newblock A solution to the learning dilemma for recurrent networks of spiking neurons.
\newblock \emph{Nature communications}, 11\penalty0 (1):\penalty0 3625, 2020.

\bibitem[Berry and Meister(1997)]{berry1997refractoriness}
Michael Berry and Markus Meister.
\newblock Refractoriness and neural precision.
\newblock \emph{Advances in neural information processing systems}, 10, 1997.

\bibitem[Bhattacharjee et~al.(2024)Bhattacharjee, Yin, Moitra, and Panda]{bhattacharjee2024snns}
Abhiroop Bhattacharjee, Ruokai Yin, Abhishek Moitra, and Priyadarshini Panda.
\newblock Are snns truly energy-efficient?—a hardware perspective.
\newblock In \emph{ICASSP 2024-2024 IEEE International Conference on Acoustics, Speech and Signal Processing (ICASSP)}, pages 13311--13315. IEEE, 2024.

\bibitem[Buzs{\'a}ki(2006)]{buzsaki2006rhythms}
Gy{\"o}rgy Buzs{\'a}ki.
\newblock \emph{Rhythms of the Brain}.
\newblock Oxford university press, 2006.

\bibitem[de~Nobel et~al.(2024)de~Nobel, Martens, Briaire, B{\"a}ck, Kononova, and Frijns]{de2024biophysics}
Jacob de Nobel, Savine~SM Martens, Jeroen~J Briaire, Thomas~HW B{\"a}ck, Anna~V Kononova, and Johan~HM Frijns.
\newblock Biophysics-inspired spike rate adaptation for computationally efficient phenomenological nerve modeling.
\newblock \emph{Hearing Research}, 447:\penalty0 109011, 2024.

\bibitem[Deng et~al.(2023)Deng, Lin, Li, and Gu]{deng2023surrogate}
Shikuang Deng, Hao Lin, Yuhang Li, and Shi Gu.
\newblock Surrogate module learning: Reduce the gradient error accumulation in training spiking neural networks.
\newblock In \emph{International Conference on Machine Learning}, pages 7645--7657. PMLR, 2023.

\bibitem[Fang et~al.(2024)Fang, Yu, Zhou, Chen, Chen, Ma, Masquelier, and Tian]{fang2024parallel}
Wei Fang, Zhaofei Yu, Zhaokun Zhou, Ding Chen, Yanqi Chen, Zhengyu Ma, Timoth{\'e}e Masquelier, and Yonghong Tian.
\newblock Parallel spiking neurons with high efficiency and ability to learn long-term dependencies.
\newblock \emph{Advances in Neural Information Processing Systems}, 36, 2024.

\bibitem[Geng and Li(2023)]{geng2023hosnn}
Hejia Geng and Peng Li.
\newblock Hosnn: Adversarially-robust homeostatic spiking neural networks with adaptive firing thresholds.
\newblock \emph{arXiv preprint arXiv:2308.10373}, 2023.

\bibitem[Gong et~al.(2013)Gong, Loi, Robinson, and Yang]{gong2013spatiotemporal}
Pulin Gong, STC Loi, Peter~A Robinson, and CYJ Yang.
\newblock Spatiotemporal pattern formation in two-dimensional neural circuits: roles of refractoriness and noise.
\newblock \emph{Biological cybernetics}, 107:\penalty0 1--13, 2013.

\bibitem[Guo et~al.(2023{\natexlab{a}})Guo, Huang, and Ma]{guo2023direct}
Yufei Guo, Xuhui Huang, and Zhe Ma.
\newblock Direct learning-based deep spiking neural networks: a review.
\newblock \emph{Frontiers in Neuroscience}, 17:\penalty0 1209795, 2023{\natexlab{a}}.

\bibitem[Guo et~al.(2023{\natexlab{b}})Guo, Zhang, Chen, Peng, Liu, Zhang, Huang, and Ma]{guo2023membrane}
Yufei Guo, Yuhan Zhang, Yuanpei Chen, Weihang Peng, Xiaode Liu, Liwen Zhang, Xuhui Huang, and Zhe Ma.
\newblock Membrane potential batch normalization for spiking neural networks.
\newblock In \emph{Proceedings of the IEEE/CVF International Conference on Computer Vision}, pages 19420--19430, 2023{\natexlab{b}}.

\bibitem[Guo et~al.(2024)Guo, Chen, Hao, Peng, Jie, Zhang, Liu, and Ma]{guo2024take}
Yufei Guo, Yuanpei Chen, Zecheng Hao, Weihang Peng, Zhou Jie, Yuhan Zhang, Xiaode Liu, and Zhe Ma.
\newblock Take a shortcut back: Mitigating the gradient vanishing for training spiking neural networks.
\newblock \emph{arXiv preprint arXiv:2401.04486}, 2024.

\bibitem[Hao et~al.(2023)Hao, Shi, Huang, Bu, Yu, and Huang]{hao2023progressive}
Zecheng Hao, Xinyu Shi, Zihan Huang, Tong Bu, Zhaofei Yu, and Tiejun Huang.
\newblock A progressive training framework for spiking neural networks with learnable multi-hierarchical model.
\newblock In \emph{The Twelfth International Conference on Learning Representations}, 2023.

\bibitem[Higuchi et~al.(2024)Higuchi, Kairat, Bohte, and Otte]{higuchi2024balanced}
Saya Higuchi, Sebastian Kairat, Sander~M Bohte, and Sebastian Otte.
\newblock Balanced resonate-and-fire neurons.
\newblock \emph{arXiv preprint arXiv:2402.14603}, 2024.

\bibitem[Hu et~al.(2024)Hu, Deng, Wu, Yao, and Li]{hu2024advancing}
Yifan Hu, Lei Deng, Yujie Wu, Man Yao, and Guoqi Li.
\newblock Advancing spiking neural networks toward deep residual learning.
\newblock \emph{IEEE Transactions on Neural Networks and Learning Systems}, 2024.

\bibitem[Huang et~al.(2024)Huang, Lin, Ren, Fu, Zhou, Liu, Pan, and Cheng]{huang2024clif}
Yulong Huang, Xiaopeng Lin, Hongwei Ren, Haotian Fu, Yue Zhou, Zunchang Liu, Biao Pan, and Bojun Cheng.
\newblock Clif: Complementary leaky integrate-and-fire neuron for spiking neural networks.
\newblock \emph{arXiv preprint arXiv:2402.04663}, 2024.

\bibitem[Ikegawa et~al.(2022)Ikegawa, Saiin, Sawada, and Natori]{ikegawa2022rethinking}
Shin-ichi Ikegawa, Ryuji Saiin, Yoshihide Sawada, and Naotake Natori.
\newblock Rethinking the role of normalization and residual blocks for spiking neural networks.
\newblock \emph{Sensors}, 22\penalty0 (8):\penalty0 2876, 2022.

\bibitem[Isaacson and Scanziani(2011)]{isaacson2011inhibition}
Jeffry~S Isaacson and Massimo Scanziani.
\newblock How inhibition shapes cortical activity.
\newblock \emph{Neuron}, 72\penalty0 (2):\penalty0 231--243, 2011.

\bibitem[Izhikevich(2007)]{izhikevich2007dynamical}
Eugene~M Izhikevich.
\newblock \emph{Dynamical systems in neuroscience}.
\newblock MIT press, 2007.

\bibitem[Jiang et~al.()Jiang, Zoonekynd, De~Masi, Gu, and Xiong]{jiangtab}
Haiyan Jiang, Vincent Zoonekynd, Giulia De~Masi, Bin Gu, and Huan Xiong.
\newblock Tab: Temporal accumulated batch normalization in spiking neural networks.
\newblock In \emph{The Twelfth International Conference on Learning Representations}.

\bibitem[Koch(2004)]{koch2004biophysics}
Christof Koch.
\newblock \emph{Biophysics of computation: information processing in single neurons}.
\newblock Oxford university press, 2004.

\bibitem[Li et~al.(2022)Li, Zhao, and Zeng]{li2022bsnn}
Yang Li, Dongcheng Zhao, and Yi Zeng.
\newblock Bsnn: Towards faster and better conversion of artificial neural networks to spiking neural networks with bistable neurons.
\newblock \emph{Frontiers in neuroscience}, 16:\penalty0 991851, 2022.

\bibitem[Lin and Huang(2024)]{lin2024spiking}
Zhanghan Lin and Haiping Huang.
\newblock Spiking mode-based neural networks.
\newblock \emph{Physical Review E}, 110\penalty0 (2):\penalty0 024306, 2024.

\bibitem[Menesse and Torres(2024)]{menesse2024information}
Gustavo Menesse and Joaqu{\'\i}n~J Torres.
\newblock Information dynamics of in silico eeg brain waves: Insights into oscillations and functions.
\newblock \emph{PLOS Computational Biology}, 20\penalty0 (9):\penalty0 e1012369, 2024.

\bibitem[Meng et~al.(2022)Meng, Xiao, Yan, Wang, Lin, and Luo]{meng2022training}
Qingyan Meng, Mingqing Xiao, Shen Yan, Yisen Wang, Zhouchen Lin, and Zhi-Quan Luo.
\newblock Training high-performance low-latency spiking neural networks by differentiation on spike representation.
\newblock In \emph{Proceedings of the IEEE/CVF conference on computer vision and pattern recognition}, pages 12444--12453, 2022.

\bibitem[Meng et~al.(2023)Meng, Xiao, Yan, Wang, Lin, and Luo]{meng2023towards}
Qingyan Meng, Mingqing Xiao, Shen Yan, Yisen Wang, Zhouchen Lin, and Zhi-Quan Luo.
\newblock Towards memory-and time-efficient backpropagation for training spiking neural networks.
\newblock In \emph{Proceedings of the IEEE/CVF International Conference on Computer Vision}, pages 6166--6176, 2023.

\bibitem[Monk and Leib(2016)]{monk2016model}
Scott Monk and Harry Leib.
\newblock A model for single neuron activity with refractory effects and spike rate estimation techniques.
\newblock \emph{IEEE Transactions on Neural Systems and Rehabilitation Engineering}, 25\penalty0 (4):\penalty0 306--322, 2016.

\bibitem[Mukhoty et~al.(2023)Mukhoty, Bojkovic, de~Vazelhes, Zhao, De~Masi, Xiong, and Gu]{mukhoty2023direct}
Bhaskar Mukhoty, Velibor Bojkovic, William de Vazelhes, Xiaohan Zhao, Giulia De~Masi, Huan Xiong, and Bin Gu.
\newblock Direct training of snn using local zeroth order method.
\newblock \emph{Advances in Neural Information Processing Systems}, 36:\penalty0 18994--19014, 2023.

\bibitem[Ratanov(2020)]{ratanov2020mean}
Nikita Ratanov.
\newblock Mean-reverting neuronal model based on two alternating patterns.
\newblock \emph{BioSystems}, 196:\penalty0 104190, 2020.

\bibitem[Rathi et~al.(2023)Rathi, Chakraborty, Kosta, Sengupta, Ankit, Panda, and Roy]{rathi2023exploring}
Nitin Rathi, Indranil Chakraborty, Adarsh Kosta, Abhronil Sengupta, Aayush Ankit, Priyadarshini Panda, and Kaushik Roy.
\newblock Exploring neuromorphic computing based on spiking neural networks: Algorithms to hardware.
\newblock \emph{ACM Computing Surveys}, 55\penalty0 (12):\penalty0 1--49, 2023.

\bibitem[Shen et~al.(2024)Shen, Ni, Xu, and Tang]{shen2024efficient}
Jiangrong Shen, Wenyao Ni, Qi Xu, and Huajin Tang.
\newblock Efficient spiking neural networks with sparse selective activation for continual learning.
\newblock In \emph{Proceedings of the AAAI Conference on Artificial Intelligence}, pages 611--619, 2024.

\bibitem[Song et~al.(2017)Song, Zhou, and Juusola]{song2017modeling}
Zhuoyi Song, Yu Zhou, and Mikko Juusola.
\newblock Modeling elucidates how refractory period can provide profound nonlinear gain control to graded potential neurons.
\newblock \emph{Physiological reports}, 5\penalty0 (11):\penalty0 e13306, 2017.

\bibitem[Taylor et~al.(2024)Taylor, King, and Harper]{taylor2024addressing}
Luke Taylor, Andrew King, and Nicol~S Harper.
\newblock Addressing the speed-accuracy simulation trade-off for adaptive spiking neurons.
\newblock \emph{Advances in Neural Information Processing Systems}, 36, 2024.

\bibitem[Vardi et~al.(2021)Vardi, Tugendhaft, Sardi, and Kanter]{vardi2021significant}
Roni Vardi, Yael Tugendhaft, Shira Sardi, and Ido Kanter.
\newblock Significant anisotropic neuronal refractory period plasticity.
\newblock \emph{Europhysics Letters}, 134\penalty0 (6):\penalty0 60007, 2021.

\bibitem[Wang et~al.(2023)Wang, Song, Wang, Xiao, Yang, Mei, and Zhang]{wang2023ssf}
Jingtao Wang, Zengjie Song, Yuxi Wang, Jun Xiao, Yuran Yang, Shuqi Mei, and Zhaoxiang Zhang.
\newblock Ssf: Accelerating training of spiking neural networks with stabilized spiking flow.
\newblock In \emph{Proceedings of the IEEE/CVF International Conference on Computer Vision}, pages 5982--5991, 2023.

\bibitem[Wu et~al.(2018)Wu, Deng, Li, Zhu, and Shi]{wu2018spatio}
Yujie Wu, Lei Deng, Guoqi Li, Jun Zhu, and Luping Shi.
\newblock Spatio-temporal backpropagation for training high-performance spiking neural networks.
\newblock \emph{Frontiers in neuroscience}, 12:\penalty0 331, 2018.

\bibitem[Xu et~al.(2023)Xu, Li, Fang, Shen, Liu, Tang, and Pan]{xu2023biologically}
Qi Xu, Yaxin Li, Xuanye Fang, Jiangrong Shen, Jian~K Liu, Huajin Tang, and Gang Pan.
\newblock Biologically inspired structure learning with reverse knowledge distillation for spiking neural networks.
\newblock \emph{arXiv preprint arXiv:2304.09500}, 2023.

\bibitem[Yao et~al.(2022)Yao, Li, Mo, and Cheng]{yao2022glif}
Xingting Yao, Fanrong Li, Zitao Mo, and Jian Cheng.
\newblock Glif: A unified gated leaky integrate-and-fire neuron for spiking neural networks.
\newblock \emph{Advances in Neural Information Processing Systems}, 35:\penalty0 32160--32171, 2022.

\bibitem[Young et~al.(2019)Young, Dean, Plank, and Rose]{young2019review}
Aaron~R Young, Mark~E Dean, James~S Plank, and Garrett~S Rose.
\newblock A review of spiking neuromorphic hardware communication systems.
\newblock \emph{IEEE Access}, 7:\penalty0 135606--135620, 2019.

\bibitem[Yu et~al.(2024)Yu, Liu, Wang, Li, and Wang]{yu2024advancing}
Chengting Yu, Lei Liu, Gaoang Wang, Erping Li, and Aili Wang.
\newblock Advancing training efficiency of deep spiking neural networks through rate-based backpropagation.
\newblock \emph{arXiv preprint arXiv:2410.11488}, 2024.

\bibitem[Yun et~al.(2019)Yun, Han, Oh, Chun, Choe, and Yoo]{yun2019cutmix}
Sangdoo Yun, Dongyoon Han, Seong~Joon Oh, Sanghyuk Chun, Junsuk Choe, and Youngjoon Yoo.
\newblock Cutmix: Regularization strategy to train strong classifiers with localizable features.
\newblock In \emph{Proceedings of the IEEE/CVF international conference on computer vision}, pages 6023--6032, 2019.

\bibitem[Zhang et~al.(2021)Zhang, Han, Niu, Gao, Chen, and Zhao]{zhang2021self}
Anguo Zhang, Ying Han, Yuzhen Niu, Yueming Gao, Zhizhang Chen, and Kai Zhao.
\newblock Self-evolutionary neuron model for fast-response spiking neural networks.
\newblock \emph{IEEE Transactions on Cognitive and Developmental Systems}, 14\penalty0 (4):\penalty0 1766--1777, 2021.

\bibitem[Zhu et~al.(2020)Zhu, Dong, Li, Huang, and Tian]{zhu2020retina}
Lin Zhu, Siwei Dong, Jianing Li, Tiejun Huang, and Yonghong Tian.
\newblock Retina-like visual image reconstruction via spiking neural model.
\newblock In \emph{Proceedings of the IEEE/CVF Conference on Computer Vision and Pattern Recognition}, pages 1438--1446, 2020.

\end{thebibliography}
}

\end{document}